\documentclass[runningheads]{llncs}
\usepackage[T1]{fontenc}
\usepackage{hyperref}
\usepackage{svg}
\usepackage{graphicx,verbatim,xcolor}
\usepackage{marvosym}

\begin{document}
\newcommand{\schuerch}{\textit{PhenoCell}}
\newcommand{\pannuke}{\textit{PanNuke}}
\newcommand{\lizard}{\textit{Lizard}}
\newcommand{\arctique}{\textit{Arctique}}
\newcommand{\unetr}{UNetR}
\newcommand{\hovernext}{HoVer-NeXt}
\newcommand{\titan}{TITAN\textsubscript{v}}
\newcommand{\provgiga}{Prov-GigaPath}
\newcommand{\phikon}{Phikon-v2}

\title{PhenoBench: A Comprehensive Benchmark for Cell Phenotyping}
%

\author{Claudia Winklmayr\textsuperscript{*,4,5}, Jérôme Lüscher\textsuperscript{*,4,5}, Nora Koreuber\textsuperscript{*,1,4,5}, Jannik Franzen\textsuperscript{*,1,3,4,5}, Fabian H. Reith\textsuperscript{*,1,2,4,5}, Elias Baumann\textsuperscript{6}, Christian M. Schürch\textsuperscript{7,8,9},
Dagmar Kainmüller\textsuperscript{\textdagger,3,4,5}, Josef Lorenz Rumberger\textsuperscript{\textdagger,2,4,5,\Letter}}  
\authorrunning{C. Winklmayr, J. Lüscher, N. Koreuber, J. Franzen, F.H. Reith, et al.}
\institute{
$^{1}$ Charité Universitätsmedizin, Berlin, Germany;
$^{2}$ Humboldt-Universität zu Berlin, Berlin, Germany;
$^{3}$ Universität Potsdam, Potsdam, Germany;
$^{4}$ Helmholtz Imaging;
$^{5}$ Max-Delbrück Center, Berlin, Germany;
$^{6}$ Institute of Tissue Medicine and Pathology, University of Bern, Bern, Switzerland;
$^{7}$ Department of Pathology and Neuropathology, University Hospital and Comprehensive Cancer Center Tübingen, Tübingen, Germany;
$^{8}$ Cluster of Excellence iFIT (EXC 2180) "Image-Guided and Functionally Instructed Tumor Therapies", University of Tübingen, Germany\\
\textsuperscript{*, \textdagger} equal contribution; when citing, it is permitted to change author order; author order was determined at random, 
\email{\textsuperscript{\Letter}\{firstnames.lastname\}@mdc-berlin.de}}  
    
\maketitle              
\begin{abstract}
Digital pathology has seen the advent of a wealth of foundational models (FM), yet to date their performance on cell phenotyping has not been benchmarked in a unified manner. We therefore propose PhenoBench: A comprehensive benchmark for cell phenotyping on Hematoxylin and Eosin (H\&E) stained histopathology images. We provide both \schuerch{}, a new H\&E dataset featuring 14 granular cell types identified by using multiplexed imaging, and ready-to-use fine-tuning and benchmarking code that allows the systematic evaluation of multiple prominent pathology FMs in terms of dense cell phenotype predictions in different generalization scenarios. We perform extensive benchmarking of existing FMs, providing insights into their generalization behavior under technical vs.\ medical domain shifts. Furthermore, while FMs achieve macro F1 scores > 0.70 on previously established benchmarks such as \lizard{} and \pannuke{}, on \schuerch{}, we observe scores as low as 0.20. This indicates a much more challenging task not captured by previous benchmarks, establishing \schuerch{} as a prime asset for future benchmarking of FMs and supervised models alike. Code and data are available on \href{https://github.com/Kainmueller-Lab/Pathology-Foundation-Model-Benchmark}{GitHub}.


\end{abstract}

\vspace{1em}
\noindent\textit{This preprint has not undergone peer review or any post-submission improvements or corrections. 
}

\section{Introduction}
\begin{figure}
\includegraphics[width=1\textwidth]{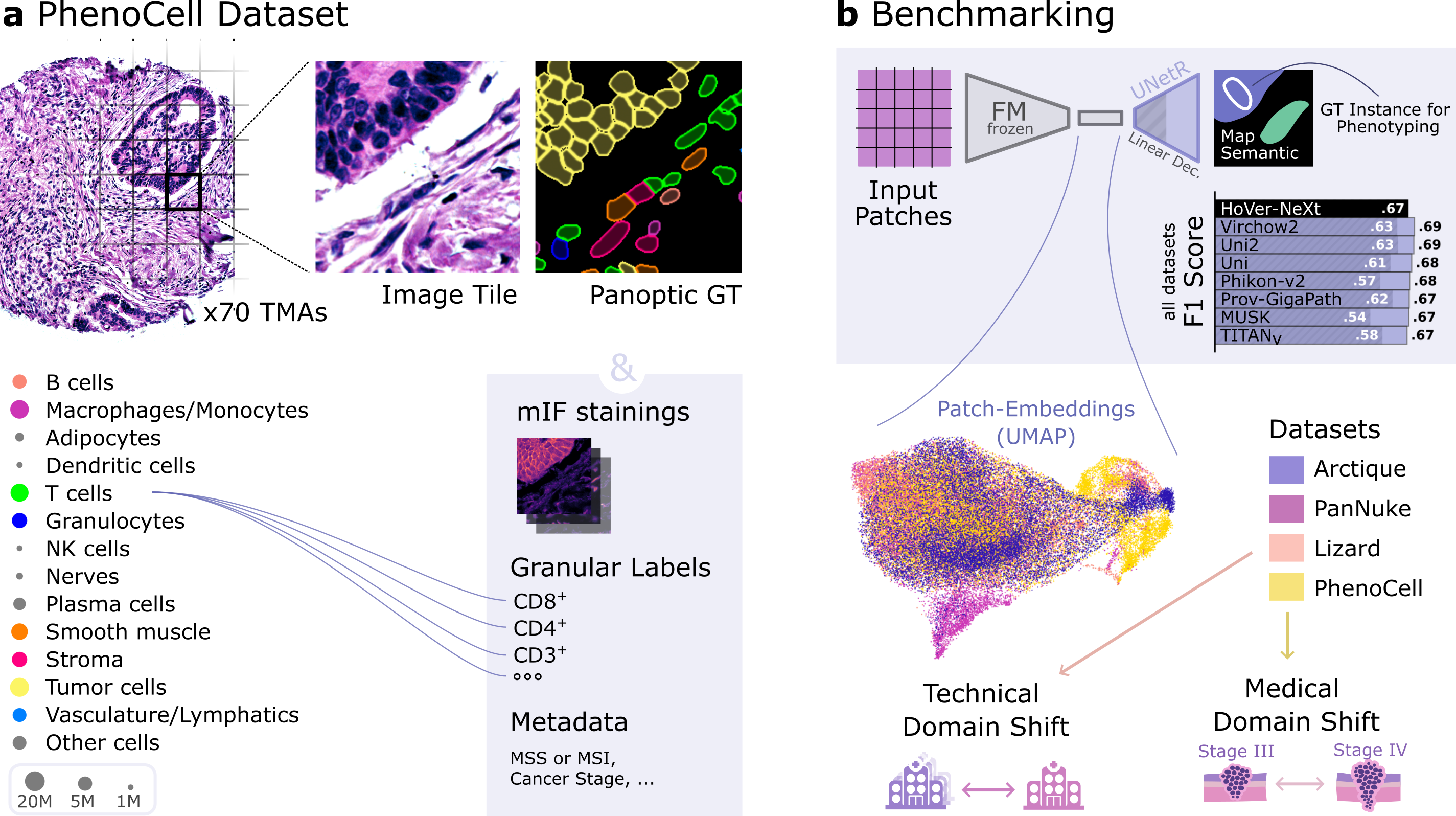}
\caption{Overview of the \schuerch{} dataset and FM benchmarking: \textbf{(a)} Example image from \schuerch{}, a zoomed-in visualization of a tile and its segmentation mask. Along the pixel-wise panoptic annotations, information such as multiplexed immunofluorescence stainings, granular labels, or patient data are provided. \textbf{(b)} Illustration of the benchmarking pipeline for evaluation of cell phenotyping capabilities of FM.}
\label{fig:graph_abstract}
\end{figure}
Automated cell type classification in Hematoxylin and Eosin (H\&E) stained histopathology images is a crucial task in computational pathology, with applications in disease diagnosis, prognosis, and treatment planning~\cite{graham2024conic}.
Recent advances in large-scale pre-trained foundation models (FMs) have enabled the development of general-purpose feature extractors demonstrating strong performance across various pathology tasks~\cite{chenGeneralpurposeFoundationModel2024,zimmermannVirchow2ScalingSelfSupervised2024,filiotPhikonv2LargePublic2024,xuWholeslideFoundationModel2024,xiangVisionLanguageFoundation2025}. However, it remains unclear to what extent these models outperform supervised baselines for cell type prediction~\cite{xu2024specialized}. Prior evaluations have been limited, either focusing on proprietary datasets with restricted access~\cite{dippel2024rudolfv} or constrained to a single dataset~\cite{chenGeneralpurposeFoundationModel2024}. Therefore, a systematic benchmark across multiple datasets is needed to determine whether pathology FMs provide an advantage over supervised baselines.

Most existing benchmarking efforts in digital pathology focus on slide- and patch-level prediction tasks~\cite{neidlinger2024benchmarking}, whereas datasets for cell phenotyping or segmentation~\cite{graham2021lizard,gamper2020pannuke} are already approaching performance saturation. Furthermore, domain shifts in prior cell-level datasets have been limited to technical factors such as differences in sample preparation, staining protocols, and scanner types~\cite{graham2021lizard}. 
However, variation may not only be of technical but also biological origin. Tumor- and molecular subtypes, as well as different stages of disease may contribute to a biological domain shift that is currently not captured  by existing benchmarks.

To bridge this gap, we publish (1) \schuerch{}, to our knowledge, the largest publicly available dataset for H\&E-based cell phenotyping with fine-grained cell type annotations (Fig. \ref{fig:graph_abstract}a), encompassing 14 distinct cell types and 88 million individual cells, much larger than all previous datasets (e.g. \lizard{}~\cite{graham2021lizard}: 6 cell types and 495k cells). 
Our dataset introduces multiple medical domain gaps, capturing variations in tumor subtypes (colon- vs. mucinous adenocarcinoma) and cancer staging (Stage 3 vs. Stage 4).
(2) We benchmark seven pathology foundation models and \hovernext{}, a strong supervised baseline, on the \schuerch{} dataset with four different dataset splits to test generalization performance under domain shifts (Fig. \ref{fig:graph_abstract}b). We further benchmark on \pannuke{}~\cite{gamper2020pannuke}, a pan-cancer dataset, \lizard{}~\cite{graham2021lizard}, a colon carcinoma dataset, and on the synthetic \arctique{} dataset~\cite{franzen2024arctique} to evaluate model generalization to a different domain.
(3) We test and compare linear probing of ViT patch-level features and a \unetr{} decoder architecture~\cite{hatamizadeh2022unetr} for dense prediction of cell phenotypes.
(4) We publicly release our benchmarking pipeline along with the dataset, enabling more comprehensive benchmarking of FMs in pathology.
\section{Datasets}

We use four datasets to evaluate the foundation model cell phenotyping performance: two established H\&E datasets, a synthetic dataset and the new \schuerch{} dataset. In selecting the datasets, we made sure that none of them was used in the training procedure of the foundation models and that each of them has unique features.

The \lizard{} dataset \cite{graham2021lizard} is one of the most widely used H\&E datasets and contains colon tissue histology images with panoptic segmentation annotations generated in a human-in-the-loop fashion. In total, \lizard{} comprises 291 fields of view (FoV) with an average size of 1,016$\times$917 pixels and a total of 495,179 individual nuclei: epithelial cells, connective tissue cells, lymphocytes, plasma cells, neutrophils, and eosinophils. 
The images in \lizard{} are acquired by six different medical centers which allows a straight-forward domain-split (\textit{Center-Split}) where training and validation samples come from five centers, while the test set consists of images from a single center (GlaS) ~\cite{rumberger2022panoptic}. 
The \pannuke{} dataset \cite{gamper2020pannuke} encompasses samples from 19 different tissue types such as colon, breast, liver, or prostate and contains a total of 205,343 nuclei annotations, generated using a human-in-the-loop data annotation pipeline. The dataset covers epithelial cells, connective tissue cells, inflammatory cells, neoplastic and dead cells. Finally, the \arctique{} dataset \cite{franzen2024arctique} is a fully synthetic dataset, constructed in a procedural fashion utilizing 3d rendering. \arctique{} replicates characteristic structures of colon histopathology images and provides exact labels for all cells. We use a subset of the full \arctique{} dataset containing 1450 FoVs of size 512$\times$512 pixel and containing a total of 487,969 cell nuclei modeled after plasma cells, lymphocytes, eosinophils, fibroblasts, and epithelial cells.

\subsection{PhenoCell Dataset}
The \schuerch{} dataset was originally published as part of a multiplexed imaging study~\cite{schurch2020coordinated} but did not include segmentation masks. The dataset was acquired using multiplexed CO-Detection by indEXing (CODEX) imaging of formalin-fixed paraffin-embedded (FFPE) tissue sections from colon carcinoma patients from the Universitätsspital Bern. Tissue samples from 35 patients were used to construct two tissue microarray (TMA) slides, each containing 70 distinct samples. These samples were imaged using 56 antibody markers across multiple cycles of staining, imaging, washing, and re-staining. Cell segmentation was performed using the \href{https://github.com/nolanlab/CODEX}{CODEX toolkit segmenter} on the DRAQ5 nuclear channel, followed by the calculation of integrated marker expressions. X-shift clustering~\cite{samusik2016automated} was applied to these marker expressions to group cells into phenotypic clusters, which were subsequently refined and merged into 28 granular cell types through manual expert review.

To create a benchmark for cell type classification in H\&E-stained histopathology, we collaborated with the authors of the original study to generate high-quality phenotyping annotations. We constructed dense annotation masks from previously unpublished segmentation data and adapted the dataset for benchmarking H\&E-based cell phenotyping. Through a rigorous quality control process, we visually inspected all segmentation masks and excluded 31 FoVs where significant false merges or false negatives were present in the cell masks. Additionally, in consultation with the original authors, we merged overly granular cell phenotypes that cannot be distinguished in H\&E images. For instance, CD3$^+$, CD4$^+$, and CD8$^+$ T cells were consolidated into a single "T cells" class. The final dataset consists of 109 high-resolution FoVs, with a total of 88 million individual cells and the following 14 distinct cell types: B cells, macrophages, nerves, dendritic cells (DC), plasma cells, granulocytes, tumor cells, T cells, stroma, adipocytes, vasculature, smooth muscle, natural killer (NK) cells, and a residual class named "other cells".

To enable detailed benchmarking under domain shifts, we provide three different dataset splits. Besides (1) \textit{Base-Split} a standard (70/15/15) train/validation/ test random split, we also provide: (2) \textit{Tumor-Type-Split} (63/16/21) where the training and validation sets contain data from patients with adenocarcinoma, while the test set consists of patients with mucinous adenocarcinoma, introducing a variation in tumor cell morphology and microenvironment.

Finally, in (3) \textit{Tumor-Stage-Split} (64/16/20) training and validation data include patients with pTNM stage 3 tumors, while the test set is composed of pTNM stage 4 tumors. This diversity in dataset splits allows for an in-depth analysis of how well FMs generalize across clinically relevant domain shifts, distinguishing this benchmark from prior datasets that split data on the medical center in which it was acquired.

\section{Pathology Foundation Models}
We benchmark seven pathology foundation models, each trained with self-super- vised (SSL) or contrastive learning on large-scale histopathology datasets. While all models utilize Vision Transformers (ViTs) as backbones, they differ in architecture, training strategies, and dataset composition.
Most models employ DINOv2~\cite{oquabDINOv2LearningRobust2023} for SSL, including Uni (ViT-L/16)~\cite{chenGeneralpurposeFoundationModel2024}, and \phikon{} (ViT-L/16)~\cite{filiotPhikonv2LargePublic2024}, both trained on a mix of publicly available and proprietary whole slide images (WSIs). Uni2 (ViT-H/14-reg8) expands upon Uni with a larger dataset and model, while Virchow2 (ViT-H/14-reg4)~\cite{zimmermannVirchow2ScalingSelfSupervised2024} applies domain-specific augmentations and regularization. 
In contrast, \provgiga{} (ViT-G/14)~\cite{xuWholeslideFoundationModel2024} introduces GigaPath, an SSL framework tailored for whole-slide representations, which the authors use for training on one of the largest proprietary WSI datasets.
Beyond DINOv2-based models, \titan{} (ViT-B/16)~\cite{dingMultimodalWholeSlide2024} is trained using iBOT~\cite{zhouImageBERTPretraining2022} and contrastive captioning~\cite{yu2022coca}, integrating vision-language alignment while supporting vision-only representations. MUSK (ViT-L/16)~\cite{xiangVisionLanguageFoundation2025} differs from all others by incorporating a BEiT3-based~\cite{wangImageForeignLanguage2022} multi-modal transformer with pathology-specific tokenization.

Another key distinction is the training dataset composition. 
While models vary in their use of proprietary and public datasets, most models rely on extensive internal datasets.
Only \phikon{} and MUSK exclusively use publicly available datasets. 
MUSK is trained on PubMed central, TCGA~\cite{CancerGenomeAtlas2022}, QUILT-1M~\cite{ikezogwoQuilt1MOneMillion2023} and PathAsst~\cite{sunPathAsstGenerativeFoundation2024}, while \phikon{} uses CPTAC \cite{edwardsCPTACDataPortal2015}, TCGA~\cite{CancerGenomeAtlas2022} and GTEx \cite{gtexconsortiumHumanGenomicsGenotypeTissue2015}. 
Uni, Uni2 and \titan{} are trained on GTEx and different subsets of proprietary data from the Mass General Brigham Hospital. For training Virchow2, proprietary data from the Memorial Sloan Kettering Cancer Center was obtained, and for \provgiga{}, data from the Providence Healthcare System was used. 

Finally, while most models extract patch-level embeddings, \provgiga{} and \titan{} focus on whole-slide representations, which may influence their ability to generalize across different histopathology tasks.
\provgiga{} uses memory-efficient dilated self-attention for generating WSI representations, whereas \titan{} uses a second transformer that takes patch-level CLS token representations from another transformer and integrates their information via the self-attention mechanism.

\begin{figure}[h]
\includegraphics[width=1\textwidth]{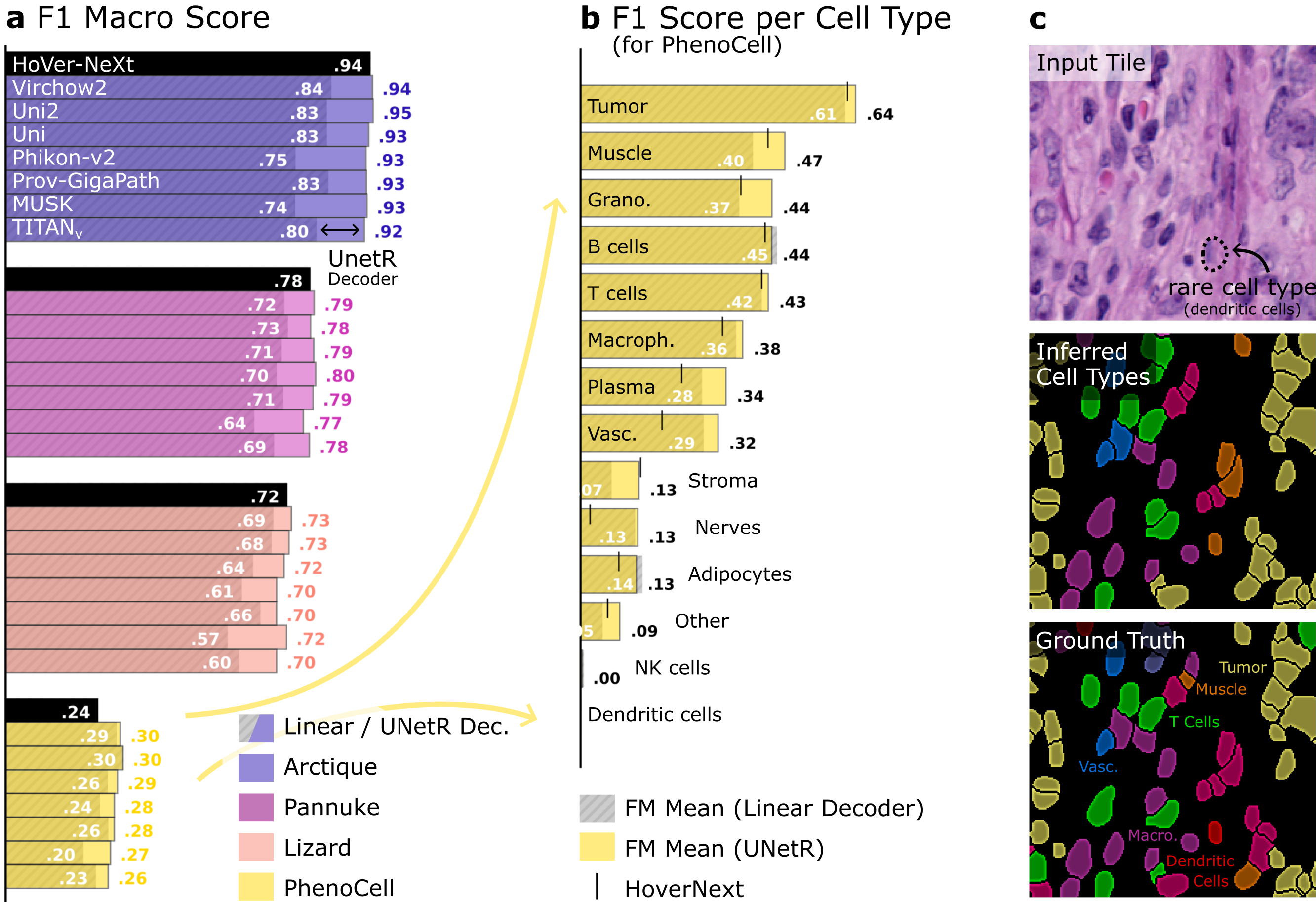}
\caption{Overview of Benchmarking results: \textbf{(a)} F1 scores of all FMs and \hovernext{} on the four datasets. Results from the linear probe are in white, \unetr{} are in dataset color. \textbf{(b)} F1 Scores per cell type for \schuerch{} dataset. \textbf{(c)} Qualitative example of our predictions and ground truth for a sample with a rare cell type (DC).}
\label{fig:fig_2}
\end{figure}

\begin{figure}
\includegraphics[width=1\textwidth]{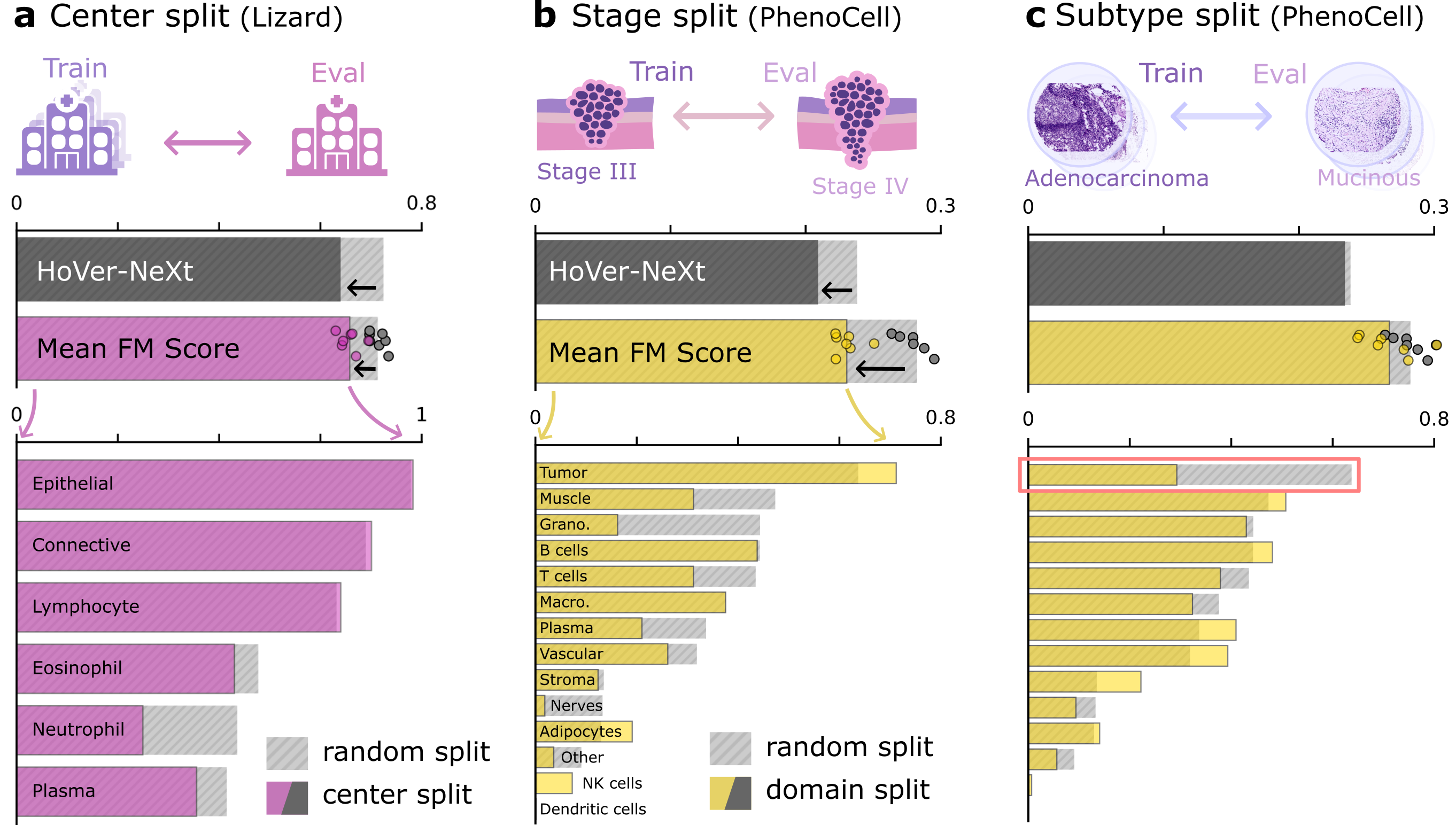}
\caption{Evaluation of FMs and \hovernext{} on medical domain splits. \textbf{(a)} \lizard{} split on medical centers for domain generalization. Here, the FMs outperform \hovernext{}. \textbf{(b)} \schuerch{} split on cancer stages 3 and 4. \textbf{(c)} \schuerch{} split on adenocarcinoma and mucinous adenocarcinoma. \textbf{Top} Mean performance on base split vs. domain split. \textbf{Bottom} Performance by cell type.}
\label{fig:domain_splits}
\end{figure}

\section{Experiments and Results}
We compare the performance of pathology foundation models with the supervised \hovernext{} baseline~\cite{baumann2024hover}, evaluating multiple aspects, including foundation model decoder architecture, dataset-specific performance, domain generalization, and data efficiency.

All experiments were conducted using Python 3.11 and PyTorch 2.5 on a NVIDIA H100 80GB GPU. 
We report the macro F1 score for all results to reflect class imbalances. The encoder weights were kept frozen for all experiments.
To assess the impact of different decoding strategies, we train models with both a simple linear projection head and the \unetr{} head~\cite{hatamizadeh2022unetr}, which incorporates a U-Net-style decoder architecture. 
Both are trained for 10k steps with early stopping. We used a learning rate of 3e\textsuperscript{-5}, with a linear warm-up of 1k steps, followed by a cosine decay. Other training parameters can be found in our code, which will be made public.

The choice of decoder has a varying effect between models and datasets (Fig. \ref{fig:fig_2}a). The greatest performance gain is observed for MUSK, where using \unetr{} improves the F1 score by up to 0.19 on the \arctique{} dataset, while the smallest effect is seen for \phikon{}, where the increase is limited to 0.03 F1 on \pannuke{}. 

Model performance varies significantly across datasets. As expected, the highest scores are achieved on the synthetic dataset \arctique{}, where models reach an average F1 of 0.93. Performance decreases for \pannuke{}: 0.79 F1, followed by \lizard: 0.71 F1, and is lowest for \schuerch{}: 0.28 F1.

To better understand model behavior on \schuerch{}, we analyze the performance across different cell types (Fig. \ref{fig:fig_2}b). The dataset is highly imbalanced, with tumor cells being the most abundant, outnumbering the least prevalent cell type, natural killer (NK) cells, by a factor of 153. As a result, classification scores are highly variable, with tumor cells achieving an F1 of 0.65, while 
NK cells score near zero. These contrasts illustrate the omnipresent challenge of recognizing rare cell types in pathology datasets~\cite{rumberger2022panoptic}.

To assess domain generalization, we evaluate models on dataset splits designed to introduce technical and biological domain gaps. In \lizard{} we use the \textit{Center-Split} which is primarily driven by variations in sample preparation, staining, and imaging.
Under these conditions, F1 for \hovernext{} drops by 0.09, whereas foundation models decrease on average by 0.06 F1, with the most robust models Uni2 and Virchow2 losing less than 0.03 F1, and the least robust model \titan{} dropping by 0.10.

In \schuerch{}, we evaluate robustness to domain shifts using the technical (Fig. \ref{fig:domain_splits}a) and medical data splits (Fig. \ref{fig:domain_splits}b). In the \textit{Tumor-Stage-Split}, foundation models exhibit a higher absolute F1 score than \hovernext{} despite experiencing a larger relative performance drop of 0.05 F1 compared to \hovernext{}’s decrease of 0.03 F1. In the \textit{Tumor-Type-Split}, \hovernext{}'s performance remains stable, decreasing by only .005 F1, whereas foundation models drop by 0.02 F1 on average. Despite this drop, foundation models still maintain a higher overall F1 of 0.27 compared to \hovernext{} with 0.23 F1 in this setting.

\begin{figure}
\includegraphics[width=\textwidth]{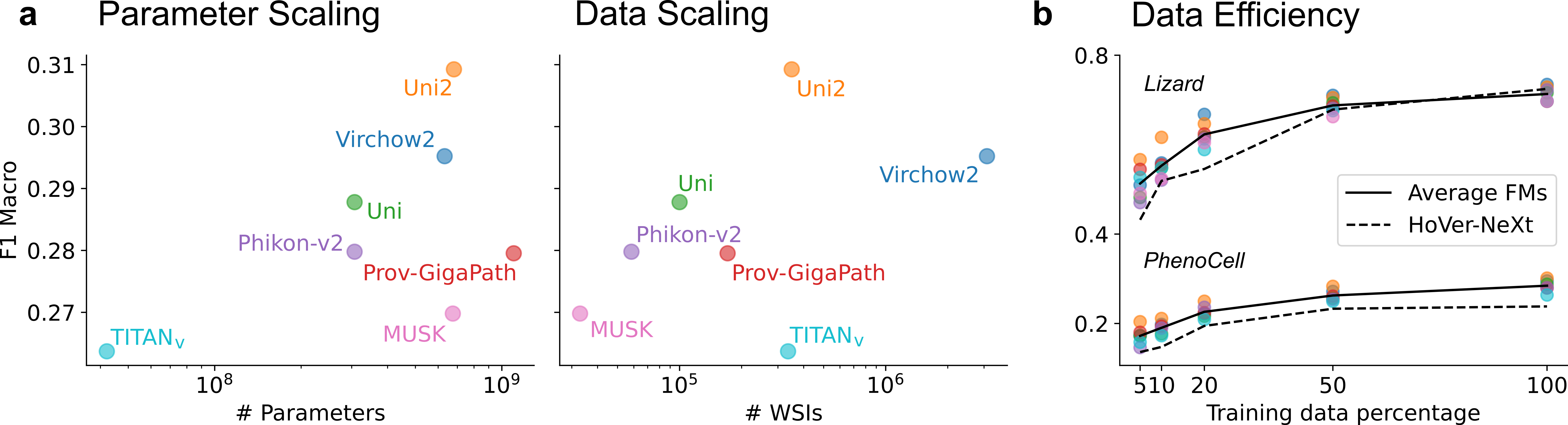}
\caption{\textbf{(a)} Performance of FMs with respect to the number of trainable parameters on the left and the number of WSIs on the right. \textbf{(b)} Performance of FMs with \unetr{} head and \hovernext{} with a reduced finetuning training set size on \lizard{} and \schuerch{}.}
\label{fig:data_eff}
\end{figure}

We also plot the performance of the FMs against their number of trainable parameters and the amount of data used for training (Fig. \ref{fig:data_eff}a). Both show a general increase with regard to their respective parameters.

To further evaluate model robustness, we conduct a data efficiency analysis (Fig. \ref{fig:data_eff}b), by progressively reducing the training set to 50, 20, 10, and 5\% of its original size while keeping the validation and test sets unchanged. On \lizard{}, foundation models show a clear advantage in low-data regimes, particularly at 5\% of the training data, where they outperform \hovernext{} by 0.08 F1, on average. This advantage is inversely proportional to the amount of training data increases. On \schuerch{}, however, the gap remains consistent across different data reductions, indicating that foundation models retain a stable advantage in settings with limited labeled data. The most data-efficient model on both \lizard{} and \schuerch{} is Uni2. On \lizard{}, Uni2 outperforms \hovernext{} by 0.14 F1 at 5\%, 0.10 F1 at 10\% and 20\%, and 0.03 ar 50\%. On \schuerch{}, the average difference of F1 between Uni2 and \hovernext{} is 0.06.

\section{Discussion and Conclusion}
Our experiments show that while FMs generally achieve competitive results in settings without domain gap (Fig. \ref{fig:fig_2}), their advantage over the \hovernext{} baseline remains limited.
The choice of decoder plays a crucial role, with the \unetr{} decoder consistently outperforming linear projection heads, due to its ability to better recover spatial details. Further improvements may be possible with more advanced decoding strategies, such as the Mask2Former decoder~\cite{cheng2022masked}.

Class imbalance remains a challenge, particularly for rare cell types, where FMs show slightly improved recognition but still struggle with morphologically ambiguous classes such as dendritic and NK cells.
Under technical domain shifts, FMs exhibit greater robustness than the supervised baseline (Fig. \ref{fig:domain_splits}a). However, FMs were more sensitive to medical domain shifts than the baseline (Fig. \ref{fig:domain_splits}b-c), albeit maintaining higher absolute performance. In low-data regimes, FMs consistently outperform the supervised baseline for the \lizard{} and \schuerch{} dataset (Fig. \ref{fig:data_eff}b). 
Among the evaluated models, approaches based on DINOv2 demonstrate strong scaling properties, while MUSK and \titan{} which were trained with alternative strategies, such as iBOT or BEiT-style masked consistency learning, perform less favorably. However, they were also trained with a comparatively small dataset (MUSK) or with less model capacity (\titan{}) and on different datasets, which limits the interpretability of observed scaling trends.

Dataset quality imposes an upper bound on performance, as both \schuerch{} and Lizard~\cite{graham2021lizard,baumann2024hover} contain label noise. The segmentation masks of \schuerch{} were created via an automatic segmentation tool, thus their quality is lower than that of other datasets that were manually or semi-manually segmented. In addition, the \schuerch{} dataset comes from a single center, thus technical variations are not reflected. Furthermore, it only contains images of colon carcinoma tissues and no other cancer types.
Finally, the reliance of many FMs on unpublished and proprietary training data limits the ability to analyze their scaling behavior.

This benchmark establishes a foundation for the evaluation of FM for cell phenotyping in digital pathology and underscores the need for more diverse high-quality data sets with precise annotations from multiple institutions. Future work should focus on expanding \schuerch{} with additional tissue types, improving annotation accuracy, and refining evaluation methodologies to better assess the robustness and generalization of foundation models in computational pathology.

\begin{credits}
\subsubsection{\ackname} Funding: DFG Research Unit DeSBi (KI-FOR 5363, project no. 459422098), (DFG) Research Training Group CompCancer (RTG2424), Synergy Unit of the Helmholtz Foundation Model Initiative, Helmholtz Einstein International Berlin Research School In Data Science (HEIBRiDS), Swiss National Science Foundation (CRSII5\_193832).

\subsubsection{\discintname}
The authors have no competing interests to declare that are
relevant to the content of this article.
\end{credits}

\bibliographystyle{splncs04}
\bibliography{references}
\end{document}